\documentclass{article}
\pdfoutput=1 
\usepackage{caption}
\usepackage{amsfonts}
\usepackage{amsmath}
\usepackage{amsthm}
\usepackage{enumitem}
\usepackage{amssymb,bm}
\usepackage[mathscr]{eucal}
\usepackage{graphicx}
\usepackage{color}
\usepackage{cite}
\usepackage{comment}
\usepackage{geometry}
\usepackage{multirow}
\usepackage{subcaption}
\usepackage{float}
\usepackage{algorithm} 
\usepackage{algorithmic} 

\usepackage{CJK}
\usepackage[utf8]{inputenc}
\usepackage{authblk} 
\title{\bf DBGSA: A Novel Data Adaptive Bregman Clustering Algorithm}
\begin{document}
\author[a]{Ying Xiao} 
\author[a]{Hou-biao Li\thanks{Corresponding author at: School of Mathematical Sciences, University of Electronic Science and Technology of China, Chengdu 611731, China.\\
E-mail address: lihoubiao0189@163.com.}
}
\author[b]{Yu-pu Zhang} 
\affil[a]{\small School of Mathematical Sciences, University of Electronic Science and Technology of China}
\affil[b]{\small School of Computer Science and  Engineering, University of Electronic Science and Technology of China}

\date{} %
\captionsetup{font={small}}
\maketitle
\begin{abstract} 
With the development of Big data technology, data analysis has become increasingly important. Traditional clustering algorithms such as K-means are highly sensitive to the initial centroid selection and perform poorly on non-convex datasets. In this paper, we address these problems by proposing a data-driven Bregman divergence parameter optimization clustering algorithm (DBGSA), which combines the Universal Gravitational Algorithm to bring similar points closer in the dataset. We construct a gravitational coefficient equation with a special property that gradually reduces the influence factor as the iteration progresses.
Furthermore, we introduce the Bregman divergence generalized power mean information loss minimization to identify cluster centers and build a hyperparameter identification optimization model, which effectively solves the problems of manual adjustment and uncertainty in the improved dataset.
Extensive experiments are conducted on four simulated datasets and six real datasets. The results demonstrate that DBGSA significantly improves the accuracy of various clustering algorithms by an average of 63.8\% compared to other similar approaches like enhanced clustering algorithms and improved datasets. Additionally, a three-dimensional grid search was established to compare the effects of different parameter values within threshold conditions, and it was discovered the parameter set provided by our model is optimal. This finding provides strong evidence of the high accuracy and robustness of the algorithm.\\

\end{abstract}

\section{Introduction}
Clustering is an essential tool in the research and application of data mining. It is an active research topic in various fields such as computer science, pattern recognition, statistics, data science, and machine learning \cite{Ezugwu}. Since the introduction of K-means, thousands of clustering algorithms have been proposed \cite{Jain}, which are based on different mechanisms, including partition-based, density-based, hierarchical, model-based, and grid-based clustering algorithms. Clustering is an unsupervised learning method that categorizes data points into multiple groups based on their similarity.
The three most commonly used clustering algorithms are hierarchical clustering, centroid-based clustering, and density-based clustering, represented by agglomerative clustering, K-means clustering, and density peak clustering, respectively. K-means, as the most widely applied clustering algorithm, is known for its simplicity and efficiency, and it is particularly suitable for spherical cluster datasets. However, K-means algorithm is highly sensitive to the initial centroid selection and outliers, leading to poor performance on non-convex datasets \cite{Ezugwu,Li,Hatamlou}. The optimization problem of K-means is to find cluster centers and assign objects to the nearest cluster center, such that the Euclidean distance between the data points and their corresponding centers is minimized to stop the iteration \cite{Cao}. Nevertheless, K-means suffers from the issue of converging to local optima.
In order to address these significant concerns related to K-means and its variations, researchers have proposed a wide range of solutions\cite{Li,Hatamlou,Kaufman}.\\
\newline
\indent Many scholars have proposed that effective distance functions play a crucial role in various machine learning and data mining techniques \cite{Wu}. Some widely used distance metrics include Euclidean, Manhattan, Chebyshev, Canberra, and other distance measures\cite{Ezugwu}. In the application of clustering algorithms, the most common distance metric is the Euclidean distance, which is a special case of the Minkowski distance. There are other distances available for specific purposes, such as the Kullback-Leibler (KL) distance, used to measure the correlation between two distributions \cite{PFraundorf}, and the Jensen-Shannon (JS) divergence, used to measure concept distance in the visual domain \cite{LWu}.
Several researchers have proposed distance metrics to improve the performance of clustering algorithms. Raeisi and Mostafa introduced a metric based on the Canberra distance, which results in larger clusters with centroids far from the origin compared to clusters with centroids closer to the origin, making it useful in cases where clusters have unequal sizes\cite{Raeisi}. However, this distance metric has considerable limitations and is not as versatile as the Bregman distance, which encompasses various useful loss functions, such as squared loss, KL divergence, logistic loss, Mahalanobis distance, Itakura Saito distance, and I-divergence \cite{Banerjee}.\\
\newline
\indent Various distance metrics proposed by scholars have shown potential in enhancing the practical applications of clustering algorithms. It has been found that hard clustering algorithms based on Bregman divergence have promising prospects. The generalization of Bregman divergence hard clustering addresses classical issues in traditional hard clustering algorithms, such as initial centroid selection, sensitivity to outliers, and the ability to find only local optima. Lei Wu and others proposed learning Bregman distance, which allows the selection of any feasible convex function to model complex distribution patterns in real datasets. Bregman distance also exhibits a strong connection with exponential families\cite{Banerjee}. When the negative entropy is chosen as the convex function, the regularized KL divergence can be represented as a special Bregman distance \cite{Wu}. Banerjee's algorithm unifies centroid-based parameter clustering methods \cite{Banerjee}, achieving the best quantization of hard clustering problems through minimizing Bregman information loss and employing mutual information loss for partition hard clustering, presenting a natural approach to solve various clustering problems based on Bregman divergence.
Br{$\acute{e}$}cheteau demonstrated that the Bregman divergence paradigm is robust against a certain level of noise, making it more suitable for clustering compared to other divergences\cite{Banerjee,Claire}. Paul proposed the power K-means\cite{Xu}, which outperforms recent techniques in the presence of outliers\cite{Paul}. Vellal optimized the same objective as Bregman hard clustering through a stepwise annealing process. Simulation-generated data showed that Bregman power K-means outperforms alternative methods on a range of exponential family data \cite{Vellal}. This algorithm retains the simplicity and scalability of classical K-means while avoiding poor local optima.
Many scholars have demonstrated the advantages of generalized Bregman distance and the effectiveness of Bregman power k-means on a range of exponential family data, as well as the rationality of initial centroid selection, from the perspective of generating datasets. Based on the aforementioned research, this paper will further extend the Bregman power k-means clustering method, constructing a loss function that minimizes Bregman information loss and identifies optimal centroid selection for hard clustering algorithms represented by K-means.\\
\newline
\indent Due to the diverse distributions of datasets analyzed using clustering algorithms, it is challenging for a given algorithm to achieve satisfactory results on all datasets. Merely enhancing the clustering algorithm has limited impact on the performance metrics\cite{Li}.
Improving clustering effectiveness involves a series of data-driven methods and optimization iterations based on natural algorithms and physical theories. Guan proposed a data-driven methods  for operational modal parameters identification, this approach takes a data-driven perspective, adapting to the distribution of the dataset and optimizing the algorithm to reduce reliance on predetermined input parameters\cite{Guan}. Over the past few decades, meta-heuristic optimization algorithms have demonstrated productivity and efficiency in addressing challenging and large-scale problems in machine learning, engineering design, and data mining applications\cite{Bala,Huang,Xie,Zhang}. There are three fundamental categories of natural heuristic algorithms: physics-based (e.g., simulated annealing), evolution-based (e.g., genetic algorithms and evolutionary algorithms), and population-based (e.g., ant colony and bee algorithms)\cite{Hegazy,Guan}.
In recent years, researchers have proposed swarm algorithms and demonstrated their effectiveness in feature selection tasks. Meta-heuristic algorithms have become a promising approach for solving clustering and classification problems. Gravity Search Algorithm (GSA) is one of the latest swarm-based meta-heuristic search techniques\cite{Dowlatshahi,Rezaei}. Krovi explored the potential feasibility of using genetic algorithms for clustering \cite{Krovi}. Krishna and Murty introduced a novel hybrid genetic algorithm\cite{Krishna}, which employs the K-Means method to globally optimize the given datasets partition into a specified number of clusters to address the expensive crossover operation problem. Hatamlou proposed a hybrid data clustering algorithm (GSA-KM) based on GSA and K-means, combining the advantages of both algorithms. GSA-KM algorithm helps K-means escape local optima and improves the convergence speed of GSA algorithm\cite{Hatamlou,Rashedi,Sun}. However, the improvement on datasets with different shapes, especially non-convex datasets is not significant, making research on enhancing dataset effects based on their distribution a current focus of attention.\\
\newline
\indent One hot topic in improving clustering results through dataset modification is the application of physics-based theories. In fact, other researchers have also explored the impact of datasets on clustering outcomes. After Wright proposed Gravity Clustering \cite{Wright}, some researchers began studying how to modify datasets to achieve better clustering performance\cite{Reichstein,Bostock,Bombarelli}. Blekas and Lagaris introduced a method called Newton Clustering \cite{Blekas}. To facilitate the determination of the number of clusters, Newton Clustering theory forces all objects to move towards cluster centers. However, Newton Clustering requires continuous iterations in data dimensions, making it unsuitable for high-dimensional datasets.
Shi proposed a Shrinkage-Based Clustering Algorithm (SBCA)\cite{Shi}. The algorithm first divides the dataset into several cells and then moves the cells along density gradient directions, efficiently detecting clusters of various densities or shapes even in noisy datasets. However, SBCA relies on cell selection without a quantitative method to compute cell sizes.
Wong introduced Herd Clustering \cite{Wong} where objects continuously tend towards the median of their neighbors. However, if the distribution of neighbors is uneven, object movements may favor sparse neighbors, significantly reducing clustering accuracy.
Qi Li presented a method based on physical analysis for improving datasets using gravitational forces \cite{Li}. They constructed a new gravitational function that brings similar objects closer and dissimilar objects farther apart. The method exhibits robustness in handling high-dimensional, Gaussian distributed, non-convex, and highly overlapping cluster datasets. The improved datasets are more suitable for many clustering algorithms. However, the method requires a large number of input parameters, and the constraints of the new gravitational formula have some limitations. Additionally, manually selecting optimal parameters within a certain range using an exhaustive method for different data distributions is time-consuming and has poor interpretation of the data.
To summarize, enhancing clustering results through dataset modification based on physics theories has become a significant area of research. Several methods have been proposed, each with its strengths and limitations, making it important to strike a balance between parameterization and performance for practical applications.\\
\newline
\indent Taking into consideration the points mentioned above, we have observed that clustering results are not only influenced by the clustering algorithms used but also by the datasets employed. Although the clustering algorithms proposed by scholars may demonstrate significant advantages under certain conditions, they also have certain limitations and may not be entirely suitable for different clustering methods and data types. Therefore, this paper aims not only to construct a Bregman generalized power mean information loss function to identify optimal initial centroids for hard clustering algorithms, such as k-means, but also to improve the datasets from a data-driven perspective. It utilizes the optimal centroids to establish an objective distance function for automatic parameter selection, constructing an optimization model to automate the identification of parameter sets. This approach provides a more robust selection of appropriate parameters, reduces uncertainty, and enhances the effectiveness of clustering methods, with k-means being a representative example.
Based on the inspiration from the above Bregman divergence hard clustering algorithm and the natural algorithms for dataset improvement, we improve the constraint conditions of gravity equation and the influence factor of gravity coefficient and propose a data-driven parameter optimization algorithm for selecting the best parameters. Furthermore, we empirically analyze the effectiveness of common Bregman divergence hard clustering algorithms. This paper aims to improve clustering effectiveness from two aspects: enhancing clustering algorithms and improving datasets. The advantages are as follows:\\
\begin{itemize}
  \item For the first time, linking the parameter selection for improved datasets to the identification of optimal initial centroids for Bregman power K-means to enhance clustering effectiveness.
  \item Optimizing the gravity-based dataset improvement algorithm and adaptive parameter selection while improving computational speed.
  \item Applicability to datasets of various shapes, with strong interpretability of the datasets.
  \item Simulating four different dataset distributions and comparing multiple distance metrics, demonstrating the effectiveness of Bregman power K-means.
\end{itemize}

\section{Related Basic Concepts}\label{sec2}
\subsection{Similarity Measurement}
\subsubsection{Euclidean Distance}\label{subsubsec1}
\indent Euclidean distance is the default distance metric used by the K-means algorithm, where $x$ and $y$ are two objects of a cluster $k$ in the cluster. Euclidean distance satisfies the three general properties of distance: non-negativity, symmetry and triangle inequality. In this paper $\|\cdot\|_2$ is $l_2$-norm, which is the Euclidean distance.
\begin{equation}
d(x, y)=\|x-y\|_2.
\end{equation}

\subsubsection{Bregman Distance}\label{subsubsec2}
\begin{bfseries}Definition 1\cite{Vellal}.\end{bfseries} Let $\varphi$ be a strictly convex function defined on a closed convex set $\Omega$, with the property of continuously differentiable real values. Arbitrary $x, y \in \Omega, \varphi: R^m \rightarrow R$ generated Bregman distance function $d_{\varphi}: R^{m \times m} \rightarrow R_{\geq 0}$ is defined as follows:
\begin{equation}
d_{\varphi}(x, y)=\varphi(\mathrm{x})-\varphi(\mathrm{y})-\langle\nabla \varphi(\mathrm{y}),(x-y)\rangle.\
\end{equation}
\\
$d_{\varphi}(x, y)$ can be considered in the $\varphi(\mathrm{x})$ and $\varphi(\mathrm{y})$ as the center of the distance between the first order approximation. It can also be described as the distance between $\varphi(\mathrm{x})$ calculated at point $x$ and the value of the tangent to $\varphi(\mathrm{y})$. For instance, take $\varphi(\mathrm{u})=\|u\|_2$ to generate the Euclidean distance. We can assume that $\varphi(\mathrm{0})=\nabla \varphi(\mathrm{0})=0$. While not necessarily symmetric and triangle inequality like the usual Euclidean distance, Bregman divergences satisfy many desirable properties that make them useful for quantifying differences. It becomes clear that Bregman distance is non-negative and remains linear.

\subsection{Gravity Search Algorithm}
\noindent Each object $x_{i j}$ of the dataset $X \in R^{n \times m}(i=1, \cdots, n ; j=1, \cdots, m)$, each row of data serves as an object $x_i=\left(x_{i 1}, x_{i 2}, \cdots, x_{i m}\right)$. According to K-Nearest Neighbor theory (KNN)\cite{Zhang,MLZhang}, each object and its the first K-Nearest objects has a gravitational pull, the total force acting on the object $i$ is
\begin{equation}
F_i=\sum_{j=1}^K F_{i j}.
\end{equation}
\indent According to Newton's second law, there is a gravitational force between any two objects in nature\cite{Hatamlou,Rashedi}. The universal gravitation search algorithm (GSA) combines the characteristics of Newton's second law, and searches not only the position of the object but also the mass of the object when solving the optimization problem. We make some new assumptions for GSA algorithm. The mutual attraction magnitude of object $i$ and $j$ at time $z$ is defined as:
\begin{equation}
F_{i j}(z)=G(z) \frac{M_i(z) \times M_j(z)}{R_{i j}(z)+\varepsilon},
\end{equation}
where $M_i(z)$ and $M_j(z)$ represent the inertial mass of object $i$ and object $j$, respectively. $\varepsilon$ is the constant that approaches zero, and $G(z)$ represents the gravitational coefficient in the $z$-th iteration. $R_{i j}$ represents the Euclidean distance between the position $x_i(z)$ and the position $x_j(z)$ corresponding to object $i$ and object $j$; The calculation formulas are as follows:
\begin{equation}
G(z)=G_0 \times e^{\left(-\frac{\alpha z}{d}\right)},
\end{equation}
\begin{equation}
R_{i j}(z)=\left\|x_i(z)-x_j(z)\right\|_2.
\end{equation}
$G_0$ is the initial value of the gravitational coefficient, $d$ represents the maximum number of iterations of $z$, and $\alpha$ is the influence factor of the gravitational coefficient. To simple the model, we assume $\alpha = 1, \varepsilon = 0.01.$

\subsection{K-Nearest Neighbors with Gravity-Based Dataset Improvement}
\indent In the gravity improved dataset experiment, $x_i$ is the $i$ dimensional vector coordinates of the original dataset, $F_{i j}^{z}$ is the $i$ and $j$ dimensional gravity formula of the improved dataset that has undergone the z-th time iteration, K-Nearest Neighbor theory redefined particles $x_i$ and its $j$ adjacent objects $o_{i j}(j \leq K)$ has undergone the first d times iteration of the gravity formula $F_{i j}$ is as follows:
\begin{equation}
F_{i j}=\left(F_{i j}^1, F_{i j}^2, \cdots, F_{i j}^d\right)=\sum_{z=1}^d\left(G^z \frac{\left\|o_{i 1}^z-x_i\right\|_2\left(o_{i j}^z-x_i\right)}{\left\|o_{i j}^z-x_i\right\|_2}\right),
\end{equation}
$o_{i 1}$ is the closest object of all the $n$ dimensional datasets, satisfy $\left\|o_{i 1}-x_i\right\|_2 \leq\left\|o_{i 2}-x_i\right\|_2 \leq \cdots \leq\left\|o_{i K}-x\right\|_2$. $K$ is the number of neighboring objects selected for $x_i$. The gravitational force on the particle $x_i$ and all its $K$ neighbors $F_i$ is
\begin{equation}
F_i=\left(F_{i 1}, F_{i 2}, \cdots, F_{i K}\right)=\sum_{z=1}^d\left(G^z \sum_{j=1}^K \frac{\left\|o_{i 1}^z-x_i\right\|_2\left(o_{i j}^z-x_i\right)}{\left\|o_{i j}^z-x_i\right\|_2}\right).
\end{equation}
\indent Based on the theory of motion of objects, the displacement formula of the stressed object is as follows:
\begin{equation}
d i s t=s=v_0 t+\frac{1}{2} a t^2=v_0 t+\frac{1}{2} \cdot \frac{F}{M} t^2.
\end{equation}
\indent To simplify the model, let the object $i$ have zero initial velocity and constant mass, we assume $v_0=0, M=1, T=\frac{1}{2} t^2$\cite{Vellal}. The effect of improving the dataset by gravity formula once is limited, and it is necessary to iterate gravity formula several times to achieve the goal of narrowing the similarity.
\begin{equation}
\triangle d i s t_i=\left(s_i^1, s_i^2, \cdots, s_i^d\right)=\sum_{z=1}^d s_i^z=x_i^{\prime}-x_i ,
\end{equation}
\begin{equation}
x_i^{\prime}=x_i+T \sum_{z=1}^d\left(G^z \sum_{j=1}^K \frac{\left\|o_{i 1}^z-x_i\right\|_2\left(o_{i j}^z-x_i\right)}{\left\|o_{i j}^z-x_i\right\|_2}\right).
\end{equation}
\indent In summary, after a series of transformations, we obtain the improved dataset $x^{\prime}=(x_1^{\prime}, x_2^{\prime}, \cdots, x_n^{\prime})=\sum_{i=1}^d\left(x_1^i, x_2^i, \cdots, x_n^i\right)$, $K, T, d$ are parameter constants.

\section{Problem Setting and Proposed Method}\label{sec3}
\subsection{Bregman Distance Extension}
This paper generates four representative clustering datasets by simulating data points based on various distribution types, and adds noise as needed. For each dataset type, clustering shape, size and noise level can be controlled by adjusting distribution parameters. Noise can be simulated by introducing randomness during data point generation or by adding some outlier values to the dataset. The following are detailed explanations of the four simulated dataset types: \\
\textbf{(1)Gaussian Distribution Dataset:} The gaussian distribution is a common data distribution used to generate clustering datasets. Mean and covariance matrices can be specified for each cluster center to control the shape and density of the dataset. \\
\textbf{(2)Binomial Distribution Dataset:} The binomial distribution can be used to generate binary datasets, where each data point can only have two possible values. Success probabilities can be specified for each cluster, and data points can be sampled from the binomial distribution. \\
\textbf{(3)Poisson Distribution Dataset:} The Poisson distribution can be used to generate datasets with discrete counts. The average count for each cluster can be specified, and data points can be sampled from the Poisson distribution. \\
\textbf{(4)Exponential Distribution Dataset:} The exponential distribution can be used to generate datasets with non-negative continuous values. The rate parameter for each cluster can be specified, and data points can be sampled from the exponential distribution.

\begin{table}[H]
\centering
\captionsetup{font={scriptsize}}
\caption{Four dataset distribution functions and their corresponding Bregman divergence}
\begin{tabular}{ccccc}
\hline
Distribution & $\varphi(\mathrm{x})$ & $d_{\varphi}(x, y)$ \\
\hline
Gaussian & $\|x\|^2$ & $\|x-y\|^2$ \\
Binomial & $\sum_{i=1}^m x_i \log x_i$ & $\sum_{i=1}^m\left[x_i \log \frac{x_i}{y_i}-\left(x_i-y_i\right)\right]$ \\
Gamma & $\alpha+\alpha \log \frac{a}{x}$ & $\frac{\alpha}{y}\left[\left(y \log \frac{y}{x}\right)+x-y\right]$ \\
Poisson& $x \log x-x$ & $x \log \frac{x}{y}-(x-y)$ \\
\hline
\end{tabular}
\end{table}

Given an m-dimensional vector space S, let $y=\left(y_1, y_2, \cdots, y_n\right)$ be a clustering centroid point and $x_i=\left(x_1, x_2, \cdots, x_3\right)$ be any data point within the cluster. The Bregman distance between $x$ and $y$ is shown in formula (2), is defined as $d_{\varphi}(x, y)=\varphi(\mathrm{x})-\varphi(\mathrm{y})-\langle\nabla \varphi(\mathrm{y}),(x-y)\rangle$. Here, $\varphi(\mathrm{\cdot})$ is a convex function that maps points in S to real numbers, $\nabla \varphi(\mathrm{y})$ represents the gradient of $\varphi$, The symbol $\langle \cdot, \cdot \rangle$ denotes the dot product between two vectors. The choice of a convex function is arbitrary, and the Bregman divergence utilizes several typical distance functions related to the distributions of the datasets, as shown in Table 1. The corresponding Bregman divergences are proven as follows:\\
\\
1) Gaussian Distance: Given $\varphi(\mathrm{x})=x^T x$, the Bregman divergence is generated as  $d_\varphi(x, y)=(x-y)^T(x-y)$, which can be considered as the classical Euclidean distance. The squared Euclidean distance is perhaps the simplest and most widely used Bregman divergence.\\
\indent Proof 1: $\nabla \varphi(y)=2 y^T, y^T x=x^T y=c$, $$\begin{aligned}
d_\varphi(x, y) & =x^T x-y^T y-\left\langle 2 y^T,(x-y)\right\rangle \\
& =x^T x-y^T y-2 y^T x+2 y^T y \\
& =x^T x-y^T y-2 y^T x+2 y^T y \\
& =x^T x-2 y^T x+y^T y \\
& =(x-y)^T(x-y).
\end{aligned}$$
2) Binomial Distance: Given $\varphi(\mathrm{x})=\sum_{i=1}^m x_i \log x_i$ generated $d_\varphi(x, y)=\sum_{i=1}^m\left[x_i \log \frac{x_i}{y_i}-\left(x_i-y_i\right)\right].$\\
\indent Proof 2: $\nabla \varphi(y)=\sum_{i=1}^m\left(1+\log y_i\right)$, $$\begin{aligned}
d_\varphi(x, y)= & \sum_{i=1}^m x_i \log x_i-\sum_{i=1}^m y_i \log y_i-\left\langle\sum_{i=1}^m\left(1+\log y_i\right),(x-y)\right\rangle \\
& =\sum_{i=1}^m x_i \log x_i-\sum_{i=1}^m y_i \log y_i-\sum_{i=1}^m\left(x_i-y_i\right)\left(1+\log y_i\right) \\
& =\sum_{i=1}^m\left[x_i \log \frac{x_i}{y_i}-\left(x_i-y_i\right)\right].
\end{aligned}$$
3) Gamma Distance: Given $\varphi(\mathrm{x})=\alpha+\alpha \log \frac{a}{x}$ generated $d_\varphi(x, y)=\frac{\alpha}{y}\left[\left(y \log \frac{y}{x}\right)+x-y\right].$\\
\indent Proof 3: $\nabla \varphi(y)=\frac{a}{y}$, $$\begin{aligned}
d_\varphi(x, y) & =\alpha+\alpha \log \frac{a}{x}-\left(\alpha+\alpha \log \frac{a}{y}\right)-\left\langle\frac{a}{y},(x-y)\right\rangle \\
& =\alpha \log \frac{y}{x}+\frac{\alpha x}{y}-\alpha \\
& =\frac{\alpha}{y}\left[\left(y \log \frac{y}{x}\right)+x-y\right].
\end{aligned}$$
4) Poisson Distance: Given $\varphi(x)=x \log x-x$ generated $d_\varphi(x, y)=x \log \frac{x}{y}-(x-y)$.\\
\indent Proof 4: $\nabla \varphi(y)=\log y $, $$\begin{aligned}
& d_\varphi(x, y)=x \log x-x-(y \log y-y)-\langle\log y,(x-y)\rangle \\
& =x \log x-x+y-x \log y \\
& =x \log \frac{x}{y}-(x-y).
\end{aligned}$$

\subsection{Four Representative Clustering Objective Functions}
 K-means and K-means power uses the traditional Euclidean distance (1), Bregman hard and Bregman power uses the Bregman distance (2). Formulas (12) - (15) represent K-means algorithm, Bregman hard clustering algorithm, K-means power algorithm and Bregman power algorithm respectively, which correspond to the minimum in-cluster variance objective function of the best center of clustering.
\begin{equation}
f_{k m e a n s}(\Theta)=\sum_{i=1}^n \min _{1 \leq j \leq k} d\left(x_i, \theta_j\right),
\end{equation}
\begin{equation}
f_{bregman-hard}(\Theta)=\sum_{i=1}^n \min _{1 \leq j \leq k} d_{\varphi}\left(x_i, \theta_j\right),
\end{equation}
\begin{equation}
f_{kmeans-power}(\Theta)=\sum_{i=1}^n \min M_s\left(d\left(x_i, \theta_1\right), d\left(x_i, \theta_2\right), \cdots, d\left(x_i, \theta_k\right)\right),
\end{equation}
\begin{equation}
f_{bregman-power}(\Theta)=\sum_{i=1}^n \min M_s\left(d_{\varphi}\left(x_i, \theta_1\right), d_{\varphi}\left(x_i, \theta_2\right), \cdots, d_{\varphi}\left(x_i, \theta_k\right)\right).
\end{equation}
where $\Theta=\left\{\theta_1, \ldots, \theta_k\right\} \subset R^m$ represents the $k$ centroids, $X=\left\{x_1, \ldots, x_n\right\} \subset R^m$ denotes the given $n$ data points. Power means are a class of generalized means defined as $M_s(y)=\frac{1}{k}\left(\sum_{i=1}^k y^s\right)^{\frac{1}{s}}$. To facilitate the optimization of the objective functions, a gradient-based approach is used. The first three objective functions use simple gradient descent. Power means has some nice properties: they are monotonic homogeneous and differentiable as follows:
\begin{equation}
\frac{\partial}{\partial y_j} M_s(\boldsymbol{y})=\left(\frac{1}{k} \sum_{i=1}^k y_i^s\right)^{\frac{1}{s}-1} \frac{1}{k} y_j^{s-1}.
\end{equation}

\indent Unlike the first three objective functions, which use simple gradient descent, $f_{bregman-power}(\Theta)$ takes advantage of this special property of Bregman divergence. First, the convexity of $\varphi$ and the properties of power functions ensure that our objective can be maximized by cutting planes. By taking a simple derivative of the generalized mean $M_s(y)$, it can be observed that the Hessian matrix of $M_s(y)$ is concave when $s \leq 1$. This provides an upper bound, offering a useful alternative function:\\
\begin{equation}
f_s(\boldsymbol{\Theta}) \leq f_s\left(\boldsymbol{\Theta}_m\right)-\sum_{i=1}^n \sum_{j=1}^k w_{m, i j} d_\phi\left(\boldsymbol{x}_i, \boldsymbol{\theta}_{m, j}\right)+\sum_{i=1}^n \sum_{j=1}^k w_{m, i j} \cdot d_\phi\left(\boldsymbol{x}_i, \boldsymbol{\theta}_j\right),
\end{equation}
\begin{equation}
w_{m, i j}=\frac{\frac{1}{k} d_\phi\left(\boldsymbol{x}_i, \boldsymbol{\theta}_{m, j}\right)^{s-1}}{\left(\frac{1}{k} \sum_{l=1}^k\left(d_\phi\left(\boldsymbol{x}_l, \boldsymbol{\theta}_{m, j}\right)^s\right)^{1-\frac{1}{s}}\right.}.
\end{equation}
$w_{m, i j}$ as the weight between $x_i$ and $\theta_j$ at the m-th iteration. \\
\indent Furthermore, the Average Minimization (MM) property from Bregman functions is applied. Lange introduced the MM principle as a general method to transform difficult optimization tasks into a series of simpler problems\cite{Lange}. It has been used for handling missing data in Maximum Likelihood Estimation (MLE) using the Expectation-Maximization (EM) algorithm as a special case.
\begin{equation}
\nabla_{\boldsymbol{\theta}_j}\left[f_s\left(\boldsymbol{\Theta}_m\right)-\sum_{i=1}^n \sum_{j=1}^k w_{m, i j} d_\phi\left(\boldsymbol{x}_i, \boldsymbol{\theta}_{m, j}\right)+\sum_{i=1}^n \sum_{j=1}^k w_{m, i j} d_\phi\left(\boldsymbol{x}_i, \boldsymbol{\theta}_j\right)\right]=0,
\end{equation}
\begin{equation}
\sum_{i=1}^n w_{m, i j} \nabla_{\boldsymbol{\theta}_j}^2 \phi\left(\boldsymbol{\theta}_j\right) \cdot\left[\boldsymbol{\theta}_j-\boldsymbol{x}_i\right]=0,
\end{equation}
\begin{equation}
\boldsymbol{\theta}_{m+1, j}=\frac{\sum_{i=1}^n w_{m, i j} \boldsymbol{x}_i}{\sum_{i=1}^n w_{m, i j}}.
\end{equation}

\indent Proofs and derivations were conducted to obtain the closed-form solution of the Expectation-Maximization (EM) equation and demonstrate its completeness\cite{Vellal}. Similar to the iterative process of updating cluster label assignments and redefining cluster means in the standard K-means algorithm, the EM equation's right-hand side is minimized to obtain ${\theta}_{m+1, j}$ for each $j$ and its corresponding weight $w_{m, i j}$, as shown in Equation (17). Formulas (18)-(21) imply a transparent and easy-to-implement approach, where a series of optimization landscapes indexed by s are implicitly explored via annealing. The resulting iterations are succinctly summarized in Algorithm 1.\\

\begin{algorithm} 
\caption{Bregman power K-means algorithm(BPK)} 
\label{alg1} 
\begin{algorithmic}[1] 
\REQUIRE dataset $X \in R^{n \times m},$ number of clusters k
\ENSURE convergent centroid cluster $\Theta=\left\{\theta_1, \ldots, \theta_k\right\} \subset R^m$ \\
\STATE Initialize $\Theta_0$
\STATE $\quad \boldsymbol{w}_{i j} \leftarrow\left(\frac{1}{k} \sum_{i=1}^k d_{\varphi}\left(x_i, \boldsymbol{\theta}_{\boldsymbol{j}}\right)^s\right)^{\frac{1}{s}-1} d_{\varphi}\left(x_i, \boldsymbol{\theta}_{\boldsymbol{j}}\right)^{s-1} \quad$
\STATE $\quad \boldsymbol{\theta}_j \leftarrow \sum_{i=1}^m \boldsymbol{w}_{i j}{ }^{-1} \sum_{i=1}^m \boldsymbol{w}_{i j}{ }^{-1} x_i \quad$
\STATE until convergence \\
\end{algorithmic}
\end{algorithm}
\indent By employing the annealing approach through the Minimization-Maximization (MM) procedure, the closed-form update algorithm inherits the objectives of Bregman hard clustering. It provides a simple and elegant solution that maintains desirable properties and interpretability while being less prone to poor local solutions. Simulation studies conducted by Vellal and others have demonstrated its superiority over alternative methods on a range of exponential family datasets\cite{Vellal},\cite{Paul}, providing a reliable theoretical foundation for our empirical analysis. Leveraging the special properties of Bregman divergence and the simple and elegant closed-form update algorithm, we further select optimal centroids, opening up possibilities for automated parameter identification in improving the dataset model.\\

\subsection{Improving Gravity Formula and Constructing Parameter Optimization Model}
As is known to all, variance can represent compactness, where a smaller variance indicates a tighter cluster. According to Lemma 1 and Theorem 1 proposed by Li and Qi\cite{Li}, the following conclusions can also be obtained. \\

\noindent \begin{bfseries}Theorem 1.\end{bfseries} Note that $T, G, K>0$, when $0<TGK<2$, the modified formula in equation (11) can make the clusters more compact. \\
\\
\textbf{Proof.}
Let $\zeta=\left\{\zeta_{i=1}^n \mid \zeta_i=\text{neighborhood }\left(\boldsymbol{x}_{\boldsymbol{i}}, K\right)=\left\{\boldsymbol{o}_{\boldsymbol{i} 1}, \boldsymbol{o}_{\boldsymbol{i} 2}, \ldots, \boldsymbol{o}_{i \boldsymbol{K}}\right\}\right\}$, where $K$ represents the given number of neighbors, and $K \ll n$. $\boldsymbol{x}_{\boldsymbol{i}}^{\prime}$ denotes the object $\boldsymbol{x}_{\boldsymbol{i}}$ after the first $d$ times iteration moving for any $\boldsymbol{o}_{\boldsymbol{i} \boldsymbol{j}} \in \zeta_i$.

$$\begin{aligned}
& \left|\boldsymbol{x}_{\boldsymbol{i}}^{\prime}-\sum_{j=1}^K \boldsymbol{o}_{\boldsymbol{i j}} / K\right| \\
& =\left|\boldsymbol{x}_{\boldsymbol{i}}+T G \sum_{j=1}^K \frac{\left\|\boldsymbol{o}_{\boldsymbol{i} 1}-x_i\right\|_2\left(\boldsymbol{o}_{\boldsymbol{i j}}-\boldsymbol{x}_{\boldsymbol{i}}\right)}{\left\|\boldsymbol{o}_{\boldsymbol{i j}}-x_i\right\|_2}-\sum_{j=1}^K \boldsymbol{o}_{\boldsymbol{i j}} / K\right| \\
& \leq \left|\boldsymbol{x}_{\boldsymbol{i}}+T G \sum_{j=1}^K\left(\boldsymbol{o}_{\boldsymbol{i j}}-\boldsymbol{x}_{\boldsymbol{i}}\right)-\sum_{j=1}^K \boldsymbol{o}_{\boldsymbol{i j}} / K\right| \\
& =\left|\boldsymbol{x}_{\boldsymbol{i}}-T G K \boldsymbol{x}_{\boldsymbol{i}}+T G \sum_{j=1}^K \boldsymbol{o}_{i j}-\sum_{j=1}^K \boldsymbol{o}_{i j} / K\right| \\
& =\left|\boldsymbol{x}_{\boldsymbol{i}}-T G K \boldsymbol{x}_{\boldsymbol{i}}+T G K \sum_{j=1}^K \boldsymbol{o}_{i j} / K-\sum_{j=1}^K {\boldsymbol{o}_{i j}} / K\right| \\
& =|1-T G K| \cdot\left|\boldsymbol{x}_{\boldsymbol{i}}-\sum_{j=1}^K \boldsymbol{o}_{i j} / K\right|.
\end{aligned}$$

Note that $T, G, K>0$, and $0<T G K<2$, then $|1-T G K|<1$, i.e.,
\begin{equation}
\left|\boldsymbol{x}_{\boldsymbol{i}}^{\prime}-\sum_{j=1}^K \boldsymbol{o}_{i j} / K\right|<\left|\boldsymbol{x}_{\boldsymbol{i}}-\sum_{j=1}^K \boldsymbol{o}_{i j} / K\right|.
\end{equation}
\indent Since $K \ll n$ and $\boldsymbol{x}_{\boldsymbol{i}} \in \zeta_i$, neighborhood $\left(\boldsymbol{x}_{\boldsymbol{i}}, K\right)$ and $\zeta_i$ are highly consistent, thus $\sum_{j=1}^K \boldsymbol{o}_{i j}/ K \approx E\left(\zeta_i\right).$ So

$$  \sum_{i=1}^K\left(\boldsymbol{x}_{\boldsymbol{i}}^{\prime}-\boldsymbol{E}\left(\zeta_i\right)\right)^2 / K<\sum_{i=1}^K\left(\boldsymbol{x}_{\boldsymbol{i}}-\boldsymbol{E}\left(\zeta_{\boldsymbol{i}}\right)\right)^2/ K.$$

Therefore, improving the dataset can reduce the variance among $ \zeta_i \in \zeta $ and bring similar data points closer together. $\blacksquare$ \\

\indent Specifically, we set $G_0=\frac{1}{N} \sum_{j=1}^N\left\|o_{i j}-x_i\right\|_2 $ in formula (5), so
$$G=\frac{1}{N} \sum_{j=1}^N\left\|o_{i j}-x_i\right\|_2 \times e^{\left(-\frac{z}{d}\right)} \leq \frac{1}{N} \sum_{j=1}^N\left\|o_{i j}-x_i\right\|_2.
$$
As the number of iterations $d$ increases, the influence of this term diminishes, adhering to the property of convergence and avoiding overfitting the dataset. Each movement results in a change in the distribution of objects. Therefore, before each movement, it is necessary to recalculate $G$ and $o_{i j}$. If $p\neq q$, then $G^p \neq G^q$ and $o_{i j}^p \neq o_{i j}^q$, $G$ is a solvable constant that decreases with the increase of iterations. For ease of representation, let $\eta=T$. Based on the above, we construct the objective function
\begin{equation}
f\left(\boldsymbol{x}_{\boldsymbol{i}}^{\boldsymbol{\prime}}\right)=\sum_{i=1}^n \min _{1 \leq j \leq k} d\left(\boldsymbol{x}_{\boldsymbol{i}}^{\boldsymbol{\prime}}, \theta_j\right).
\end{equation}

\indent Assuming $0<T, TGK<1, 1\leq d, K \leq 10$, the iterative step sizes for the three parameters are: $\Delta \eta = 0.01, \Delta d=\Delta K=1.$ Let $g\left(\boldsymbol{x}_{\boldsymbol{i}}^{\boldsymbol{\prime}}\right)=\Delta x_i=\sum_{i=1}^d\left(G^z \sum_{j=1}^K \frac{\left\|o_{i 1}^z-x_i\right\|_2\left(o_{i j}^z-x\right)}{\left\|o_{i j}^z-x_i\right\|_2}\right)$ in formula (11), we can obtain
\begin{equation}
\boldsymbol{x}_{\boldsymbol{i}}^{\boldsymbol{\prime}}=x_i+\eta * g\left(\boldsymbol{x}_{\boldsymbol{i}}^{\mathbf{\prime}}\right).
\end{equation}
The details as shown in Algorithm 2.

\begin{algorithm} 
\caption{Data-Driven Adaptive Bregman Clustering Algorithm (DBGSA)} 
\label{alg2} 
\begin{algorithmic}[1] 
\REQUIRE dataset $X \in R^{n \times m}$, centroid cluster $\Theta=\left\{\theta_1, \ldots, \theta_k\right\} \subset R^m$\\
\ENSURE impove dataset $X^{\prime}\in R^{n \times m}$ \\
\STATE Initialize parameter set $\left(\eta_0, K_0, d_0\right)=(1,1,1)$ and function value $f\left(x_{\boldsymbol{i}}\right)$
\STATE \quad $K,d \leftarrow uniform(1,11,1)$
\STATE \quad $\eta \leftarrow \eta_0- Kd*\Delta \eta$
\STATE \quad \textbf{for} $i$ \textbf{in} range ($n$) \textbf{do}
\STATE \quad \quad $\boldsymbol{x}_{\boldsymbol{i}}^{\boldsymbol{\prime}}\leftarrow x_i+\eta * g\left(\boldsymbol{x}_{\boldsymbol{i}}^{\mathbf{\prime}}\right)$\\
\STATE \quad \textbf{end for}\\
\STATE \quad  \textbf{until} formula (23) is minimum or no longer changes\\
\end{algorithmic}
\end{algorithm}


\indent Algorithm 2 derives the optimal parameters by constructing an adaptive parameter model, and using these optimal parameters in the gravity formula significantly improves the clustering effectiveness of the dataset. DBGSA proposed in this paper follows these ideas: First, improve the dataset by combining K-nearest neighbors and the gravity formula, transforming a bad dataset into a good one. Then, it introduces an adaptive parameter model for optimizing the gravity algorithm's initial centroids, improving the computational speed. Finally, four classic dataset distribution types are used to compare multiple distance metrics and demonstrate the applicability of Bregman power K-means to various datasets, see Experiment 2.


\section{Empirical Performance and Results}\label{sec4}
To verify the applicability and accuracy of different clustering algorithms in different scenarios, we conducted experiments on six different real datasets in Table 2.\\
\indent (1) Iris dataset: This dataset contains 150 records with 4 features.\\
\indent (2) Digit dataset: This dataset contains a total of 1797 records with 64 features. \\
\indent (3) Seeds dataset: This dataset contains a total of 210 records and has 7 features.\\
\indent (4) Wine dataset: This dataset contains a total of 178 records with 13 characteristics.\\
\indent (5) Breast dataset: This dataset contains a total of 569 records with 30 features. \\
\indent (6) Wireless dataset: This dataset contains a total of 2000 records and has 7 characteristics.
\begin{table}[H]
\centering
\captionsetup{font={scriptsize}}
\caption{Six representative real datasets.}
\begin{tabular}{ccccc}
\hline
Dataset& Number& Dimension& Category& Size \\
\hline
Iris& 150& 4& 3& 600\\
Seeds& 210& 7& 3& 1470\\
Wine& 178& 13& 3& 2314\\
Wireless& 2000& 7& 4& 4000\\
Breast& 569& 30& 2& 17070\\
Digit& 1797& 64& 10& 115008\\
\hline
\end{tabular}
\end{table}

To evaluate the clustering results, we used two representative performance metrics: Adjusted Rand Index (ARI) and Normalized Mutual Information (NMI). ARI addresses the limitation of Rand Index (RI) in describing the similarity of randomly assigned cluster labels. ARI's values range from -1 to 1, where higher values close to 1 indicate higher similarity between the predicted and true cluster labels, values close to 0 suggest random assignment of clusters, and negative values indicate poor predictions of cluster labels.NMI measures the similarity between predicted and true cluster labels based on mutual information scores, and it normalizes the MI values between 0 and 1, enabling comparisons across different datasets. Higher values close to 1 indicate higher similarity, while values close to 0 indicate random assignment of clusters.By using these performance metrics, we can assess and compare the clustering results obtained from different algorithms on various datasets.\\
\\
\noindent \textbf{Experiment 1: Data Visualization and Data Preprocessing Techniques}\\
Common data preprocessing techniques include feature scaling and feature dimensionality reduction. Standardization and normalization are common methods for feature scaling, which transform the data into specific distributions or ranges. Principal Component Analysis (PCA) and t-distributed Stochastic Neighbor Embedding (TSNE) are common methods for feature dimensionality reduction, which address dimensionality redundancy and multicollinearity problems. Below are their advantages, disadvantages, and expressions: \\
\indent 1) \textbf{Standardization:} It eliminates the scale difference between features, improves the model's robustness to outliers, and is applicable to various models. However, it incurs partial information loss and sensitivity to outliers. Standardization is achieved by subtracting the mean and dividing by the standard deviation of the original data, as shown in the following expression: \\
$$x_{new}=(x-\operatorname{mean}(x)) / \operatorname{std}(x).$$
\indent 2) \textbf{Normalization:} It scales the data to a specified range of [0, 1] or [-1, 1], making it easier for the model to learn appropriate weight parameters and improve convergence speed. However, it incurs partial information loss and sensitivity to outliers. Normalization is achieved by linearly transforming the data to the specified range, as shown in the following expression:\\
$$ x_{new}=(x-\min (x)) /(\max (x)-\min (x)).$$
\indent 3) \textbf{PCA:} It can eliminate the factors that influence each other among the original data, achieving dimensionality reduction while retaining most of the important information. However, it incurs partial information loss and relatively poor interpretability. In PCA, principal components are obtained by linearly combining the original variables. These new variables are arranged in decreasing order of their variances, and the contribution of each variable can be measured by the variance explained ratio in the principal components. \\
\indent 4) \textbf{TSNE:} It mainly performs nonlinear dimensionality reduction, which has the advantage of selecting similar objects with higher probability and dissimilar objects with lower probability. It maps multidimensional data to two or more dimensions suitable for observation, preserving local structure and better displaying data structure in low-dimensional space. However, it has higher computational complexity.\\
\\
\indent The choice of normalization, standardization, and dimensionality reduction methods should be based on the specific characteristics of different datasets. If the data distribution is known and close to normal distribution, and the model requires feature absolute value magnitudes, standardization can be considered. If the data distribution is unknown or does not conform to normal distribution, and only the relative weights between features need to be considered, normalization can be considered. If there are problems of dimensionality redundancy and multi-collinearity, PCA and t-SNE can be considered. Additionally, the clustering effect can be analyzed through experiments and adjustments based on the model requirements and actual results, and visualization tools can be used to choose the most suitable preprocessing method. As shown in Figures 1 and 2, Iris dataset is suitable for normalization, Wine dataset is suitable for standardization, and high-dimensional datasets require dimensionality reduction. The preprocessing of the four representative datasets shows good visualization results.\\

\begin{figure}[H]
\centering
\begin{minipage}{0.48\linewidth}
\centering
\includegraphics[width=\linewidth]{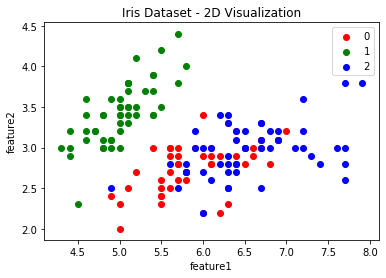}
\end{minipage}
\vspace{3pt}
\begin{minipage}{0.48\linewidth}
	\includegraphics[width=\linewidth]{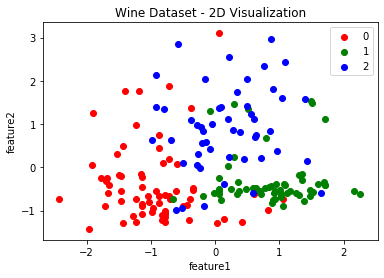}
\end{minipage}
\captionsetup{font={scriptsize}}
\caption{Visualization of the Iris and Wine dataset after feature scaling.}
\end{figure}
\begin{figure}[H]
\begin{minipage}{0.48\linewidth}
\centering
\includegraphics[width=\linewidth]{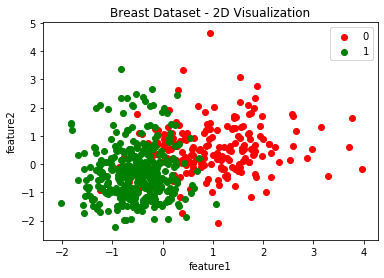}
\end{minipage}
\begin{minipage}{0.48\linewidth}
	\includegraphics[width=\linewidth]{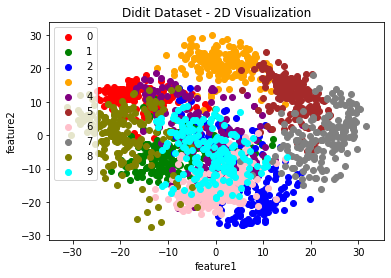}
\end{minipage}
\captionsetup{font={scriptsize}}
\caption{Visualization of Breast and Digit dataset after feature dimensionality reduction}
\end{figure}

\noindent \textbf{Experiment 2: Effectiveness Of Bregman Power K-means Algorithm in Selecting Optimal Centroids.}\\
To validate the effectiveness of the Bregman power K-means algorithm, we first consider simulating datasets from various exponential families in the plane. We randomly generate Gaussian, binomial, Poisson, and gamma distributions with true centers at (10, 10), (20, 20), and (40, 40) in $R^2$, add some noise to the generated dataset. Assuming there are 99 samples in each distribution, we repeat the experiment 250 times to obtain the mean and variance of various distance metrics. In this experiment, we use the adjusted Rand index (ARI) as the evaluation metric.
\begin{table}[H]
\centering
\captionsetup{font={scriptsize}}
\caption{Generates the ARI mean and standard deviation of random data over Gaussian, binomial, Poisson, and gamma distributions}
\begin{tabular}{ccccc}
\hline
Methods& Gaussion& Binomial& Possion& Gamma \\
\hline
K-means&0.808$\pm$0.012& 0.865$\pm$0.012& 0.722$\pm$0.014& 0.484$\pm$0.009 \\
Bregman Hard& 0.837$\pm$0.012& 0.886$\pm$0.011& 0.882$\pm$0.010& 0.868$\pm$0.005 \\
K-means Power& \textbf{0.927$\pm$ 0.003}& 0.915$\pm$0.004& 0.888$\pm$0.006& 0.677$\pm$0.008 \\
Bregman power& \textbf{0.927$\pm$0.003}& \textbf{0.961$\pm$0.003}& \textbf{0.916$\pm$0.004}& \textbf{0.879$\pm$0.004} \\
\hline
\end{tabular}
\end{table}

\indent Bregman divergence can provide an ideal distance metric for the generated clustered data under their respective exponential families. We apply K-means, Bregman hard clustering, K-means power and Bregman power K-means algorithm to these four dataset distribution settings respectively. We randomly select initial centroids based on the uniform distribution range of all data points. The K-means power and Bregman power K-means use an initial power factor $s_0=-0.2$. From Table 3, we observe that the Bregman power K-means algorithm always achieves the best performance in different exponential family data distributions when compared with other algorithms. The mean ARI values are generally greater than 0.9, and the standard deviation is not higher than 0.005, indicating good robustness and effectiveness. Next, we will use the Bregman power k-means algorithm to find the optimal centroids for the parameter optimization experiment of improving the dataset.\\

\noindent \textbf{Experiment 3: The Effectiveness of DBGSA Algorithm }\\
To validate the effectiveness of DBGSA, we compare it with the classical algorithms Herd, SBCA, and HIBOG, which use improved datasets to enhance clustering performance. We use NMI as the metric and select three representative clustering algorithms for comparison: Agglomerative, K-means and Peak representing hierarchical clustering, centroid-based clustering, and density-based clustering, respectively.
\begin{table}[H]
	\centering
	\captionsetup{margin=1cm}
	\captionsetup{font={scriptsize}}
	\caption{Comparisons of Herd, SBCA, HIBOG and POGDA with NMI values of the original algorithm.}
	\begin{tabular}{cccccccc}
		\hline
		Methods& Dataset& K-means& Agglomerative& Peak \\
		\hline
  \multirow{6}*{real} &Breast &0.422 &0.261 &0.166 \\
 &Digit &0.738 &0.856 &0.716 \\
 &Iris &0.748 &0.758 &0.707 \\
 &Seeds &0.691 &0.724 &0.706 \\
 &Wine &0.423 &0.410 &0.384 \\
 &Wireless &0.885 &0.906 &0.864 \\
		\hline			
  \multirow{6}*{Herd} &Breast &0.611 &0.677 &	0.408 & \\
 &Digit &0.740 &	0.858 &	0.781 & \\
 &Iris &0.752 &	0.750 &	0.778 & \\
 &Seeds &0.722 &	0.699 &	0.705 & \\
  &Wine &0.847 &	\textbf{0.907} &	0.697 & \\
 &Wireless &0.885 &	0.862 &	0.80 & \\
		\hline		
  \multirow{6}*{SBCA} &Breast cancer &0.611 &0.497 &0.454 \\
 &Digit &0.740 &0.849 &0.639 \\
 &Iris &0.748 &0.786 &0.883 \\
 &Seeds &0.730 &0.75 &0.739 \\
 &Wine &0.874 &\textbf{0.907} &0.646 \\
 &Wireless &0.829 &0.883 &0.867 \\
		\hline			
  \multirow{6}*{HIBOG} &Breast &0.705 &0.708 &0.502 \\
 &Digit &0.882 &0.877 &\textbf{0.915} \\
 &Iris &0.813 &0.803 &0.793 \\
 &Seeds &0.772 &0.798 &0.726 \\
 &Wine &0.889 &0.874 &0.863 \\
 &Wireless &0.854 &0.878 &0.923 \\
		\hline			
  \multirow{6}*{DBGSA} &Breast &\textbf{0.764} &\textbf{0.781} &\textbf{0.701} \\
 &Digit &\textbf{0.911} &\textbf{0.902} &0.913  \\
 &Iris &\textbf{0.931} &\textbf{0.900} &\textbf{0.949}  \\
 &Seeds &\textbf{0.801} &\textbf{0.803} &\textbf{0.791}  \\
 &Wine &\textbf{0.909} &0.877 &\textbf{0.882}  \\
 &Wireless &\textbf{0.922}  &\textbf{0.932} &\textbf{0.924}  \\

		\hline
	\end{tabular}
\end{table}
The experimental results show that our proposed improved clustering algorithms have better accuracy and robustness for different datasets. Different clustering algorithms have significant differences in their applicability and accuracy. In the Wine dataset, Herd and SBCA perform best with the Agglomerative algorithm, while in other datasets, they perform poorly. HIBOG performs better than Herd and SBCA overall, but our proposed BPGDA algorithm has higher performance metrics than HIBOG. Moreover, it builds a data-driven model that automatically identifies multiple parameters, reduces uncertainty and randomness, and provides better interpretability.
\begin{table}[H]
	\centering
	\captionsetup{margin=1cm}
	\captionsetup{font={scriptsize}}
	\caption{Comparisons of Herd, SBCA, HIBOG and POGDA with NMI incremental values of the original algorithm.}
	\begin{tabular}{cccccccc}
		\hline
		Methods& Dataset& K-means& Agglomerative& Peak \\
		\hline
  \multirow{7}*{Herd} &Breast &44.8\% &159.4\% &145.8\% \\
 &Digit &0.3\% &0.2\% &9.1\% \\
 &Iris &0.5\% &-1.1\% &10.0\% \\
 &Seeds &4.5\% &-3.5\% &-0.1\% \\
 &Wine &100.2\% &121.2\% &81.5\% \\
 &Wireless &0.0\% &4.9\% &-7.4\% \\
 &Average &\textbf{25.1\%} &\textbf{45.2\%} &\textbf{39.8\%} \\
		\hline		
  \multirow{7}*{SCAN} &Breast &44.8\% &90.4\% &173.5\% \\
 &Digit &0.3\% &-0.8\% &-10.8\% \\
 &Iris &0.0\% &3.7\% &24.9\% \\
 &Seeds &5.6\% &3.6\% &4.7\% \\
 &Wine &106.6\% &121.2\% &68.2\% \\
 &Wireless &-6.3\% &-2.5\% &0.3\% \\
 &Average &\textbf{25.2\%} &\textbf{35.9\%} &\textbf{43.5\%} \\
		\hline		
  \multirow{7}*{HIBOG} &Breast &67.1\% &171.3\% &202.4\% \\
 &Digit &19.5\% &2.5\% &27.8\% \\
 &Iris &8.7\% &5.9\% &12.2\% \\
 &Seeds &11.7\% &10.2\% &2.8\% \\
 &Wine &110.2\% &113.2\% &124.7\% \\
 &Wireless &-3.5\% &-3.1\% &6.8\% \\
 &Average &\textbf{35.6\%} &\textbf{50.0\%} &\textbf{62.8\%} \\
		\hline		
  \multirow{7}*{DBGSA} &Breast &80.9\% &199.3\% &322.3\% \\
 &Digit &23.4\% &5.4\% &27.5\% \\
 &Iris &24.4\% &18.7\% &34.2\% \\
 &Seeds &15.9\% &11.0\% &12.0\% \\
 &Wine &114.8\% &113.9\% &129.7\% \\
 &Wireless &4.2\% &2.9\% &6.9\% \\
 &Average &\textbf{44.0\%} &\textbf{58.5\%} &\textbf{88.8\%} \\
		\hline
	\end{tabular}
\end{table}

\indent As a more obvious comparison, Table 4 shows the NMI increment value of Herd, SBCA, HIBOG, and DBGSA compared to the original algorithms. It can be observed that the DBGSA algorithm performs best among the three representative K-means algorithms, Agglomerative algorithm, and Peak algorithm. Compared to the original algorithms, it achieves an average improvement of 44.0\%, 58.5\%, and 88.8\% for the six real datasets. Overall, DBGSA improves by 63.8\%, HIBOG improves by 49.5\%, Herd improves by 36.7\%, and SBCA improves by 34.9\%.

\begin{figure}[H]
	\centering
	\begin{minipage}{0.32\linewidth}
		\centering
		\includegraphics[width=\linewidth]{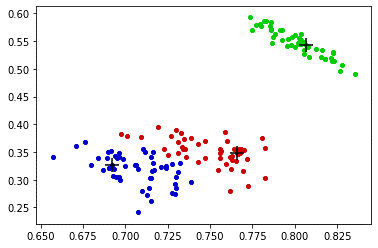}
	\end{minipage}
	\vspace{3pt}
	\begin{minipage}{0.32\linewidth}
		\includegraphics[width=\linewidth]{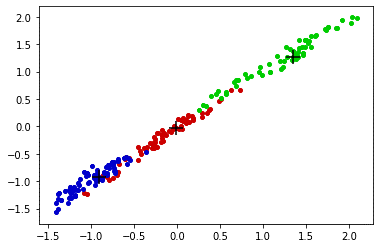}
	\end{minipage}
	\vspace{3pt}
	\begin{minipage}{0.32\linewidth}
		\includegraphics[width=\linewidth]{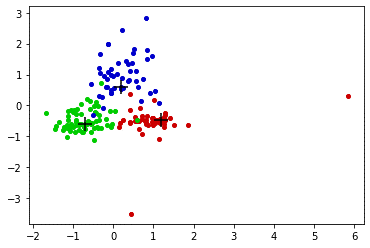}
	\end{minipage}	
	\vfill
	\begin{minipage}{0.32\linewidth}
		\centering
		\includegraphics[width=\linewidth]{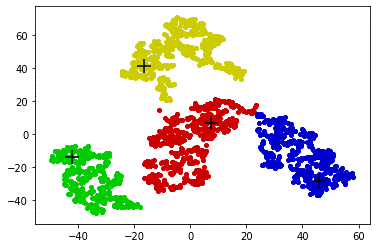}
	\end{minipage}
	\vspace{3pt}
	\begin{minipage}{0.32\linewidth}
		\includegraphics[width=\linewidth]{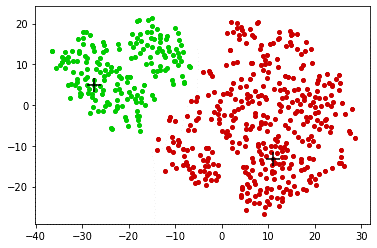}
	\end{minipage}
	\vspace{3pt}
	\begin{minipage}{0.32\linewidth}
		\includegraphics[width=\linewidth]{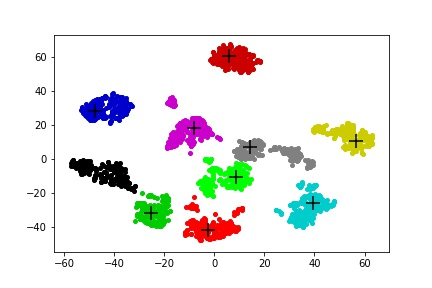}
	\end{minipage}
	\vspace{3pt}
	\captionsetup{font={scriptsize}}
	\caption{Visualization of 6 representative data sets after improved DBGSA algorithm.}
\end{figure}
Figure 3 shows the visualization of the improved datasets based on data-driven improvements and Bregman divergence parameter optimization model (DBGSA) using the density-based clustering algorithm (Peak) as an example. Corresponding to Iris, Wine, Seeds, Breast, Wireless, and Digit datasets, DBGSA effectively classifies these six representative datasets and can be widely used in practical clustering analysis problems.

\section{Conclusions}\label{sec5}
Based on four simulated exponential distribution datasets and four typical distance metric algorithms, the Bregman power K-means divergence provide the best performance in finding centroids. Compared to three representative clustering algorithms and Herd, SCAN, HIBOG algorithms for improving datasets, the proposed data-driven Bregman divergence parameter optimization algorithm (DBGSA) shows the better performance, with the higher improvement in all types of clustering algorithms, averaging 63.8\%. The proposed Bregman divergence centroid parameter optimization model can be widely used for clustering-related hyperparameter identification and improved dataset automatic parameter adjustment. Due to space limitations, our experiments only selected representative metrics ARI and NMI. In the future, we will continue to study relevant theories and improve the experimental section.


\end{document}